\documentclass{sage}

\usepackage[utf8]{inputenc} % set input encoding (not needed with XeLaTeX)

\usepackage{geometry} % to change the page dimensions
\usepackage{graphicx} % support the \includegraphics command and options
\graphicspath{{../../media/rsmg/img/}{../../media/thesis/img/}}

\usepackage[parfill]{parskip} % Activate to begin paragraphs with an empty line rather than an indent
\usepackage[british]{babel}
\usepackage{booktabs} % for much better looking tables
\usepackage{array} % for better arrays (eg matrices) in maths
\usepackage{paralist} % very flexible & customisable lists (eg. enumerate/itemize, etc.)
\usepackage{verbatim} % adds environment for commenting out blocks of text & for better verbatim
\usepackage{subfig} % make it possible to include more than one captioned figure/table in a single float
\usepackage{hyperref} % make cliccable the table of contents
\usepackage{fancyhdr} % This should be set AFTER setting up the page geometry
\usepackage[nottoc,notlof,notlot]{tocbibind} % Put the bibliography in the ToC
\usepackage[titles,subfigure]{tocloft} % Alter the style of the Table of Contents

\usepackage{setspace}
\usepackage{algorithm}
\usepackage{algorithmic}
\floatname{algorithm}{Algorithm}

\newcommand{\newdef}[1]{\textbf{#1}}

\usepackage{listings}

\title{\large{University of Birmingham\\
	School of Computer Science}\\
	\vspace{8mm}
	\Large{\textsc{Grasping and Manipulation with a Multi-Fingered Hand}} \\
	\LARGE{\textbf{RSMG Report: Thesis Proposal}}}
\author{Claudio \textsc{Zito} \\
	\vspace{2mm}\\
	\emph{Supervisors:} Prof. Jeremy \textsc{Wyatt}, Dr. Rustam \textsc{Stolkin}\\
	\emph{Thesis group members:} Dr. Richard \textsc{Dearden}, Dr. Hamid \textsc{Dehghani}
	\vspace{4mm}}
\date{\today} % Activate to display a given date or no date (if empty),
\begin{document}
\maketitle

\pagenumbering{roman} \tableofcontents \newpage \pagenumbering{arabic}

%------------------------------------------------------------------------------------------------------------------------------------------------------------------------------------------------
%	INTRODUCTION
%------------------------------------------------------------------------------------------------------------------------------------------------------------------------------------------------
\section{Introduction}

In this thesis, we will attempt to develop new algorithms that enable an agent to deal with uncertainty in application to planning of robot grasps and manipulative actions. We focus on constructing a system where an agent interacts with its surrounding environment by selecting the most appropriate action at any time in a task-oriented fashion.

Motion planning techniques are common tools to solve problems where the search occurs directly in the configuration space of the object to be moved (see e.g. Rapidly-exploring Random Tree (RRT)~\cite{bib:lavalle_2001}). Motion planning was originally concerned with problems such as how to move a piano from one room to another of a house without colliding with obstacles or bounding walls. In manipulation and grasping, however, the object to be moved is indirectly controlled by contact with a robot manipulator. This is a completely different problem which requires the ability to search for a sequence of manipulative actions that will move the object from a start to a goal configuration, e.g. the start configuration might be a bottle sitting on a table and the goal configuration might be the same bottle grasped by the agent and tilted to pour out the content. 

Our main innovation is to split the planning problem into: (1) an RRT-based planner operating in the configuration space of the object, and (2) a local planner which generates a sequence of actions in the joint space that will move the object between two pairs of nodes in the RRT.   

The simplest version of this problem is one in which the robot can manipulate the object by a single contact point. In this case, the robot is equipped with a single finger, and the set of manipulative actions is constrained only to push operations. We have shown that this two-level strategy enables us to find successful pushing plans in a simulated environment. The local push planner uses a randomised depth-first search procedure for finding locally appropriate sequences of pushes to reach from a previous node towards the next candidate node suggested by the RRT planner.

We aim to extend the local planner in order to deal with a set of manipulative actions which involve more than one contact (e.g. grasping). This is not a trivial extension for several reasons:
\begin{description}
\item[Physics simulation with multiple contacts] the local push planner defines its action-selection behaviour making use of a physics simulator for prediction, i.e. the physics simulator provides a prediction of the next object configuration which will result given the current configuration and the selected action. Unfortunately our recent researches have shown that the most popular physics simulators are not stable in prediction when multiple forces are applied to the same object. This means we need to figure out a different model to predict actions' outcomes. A possible way might be using a different kind of forward model, as e.g. learned predictors (see Section~\ref{sec:forwardmodel}). Alternately we might setup a static real scenario where testing grasp stability;

\item[Grasp quality] the local push planner when executes the selected actions, behaves as a closed-loop controller and checks whether or not the actual object's configuration state is decreasing a distance function from the previous (checked) object's configuration and the candidate node selected by the RRT planner. In this simple case, the state space represents only the object pose in a 3D space; and the metric distance denotes the distance between two configuration states in terms of rotational and transitional displacement. In grasping, however, the state space should also include the robot manipulator configuration; and the evaluation function should denote a qualitative measurement of the grasp state~\cite{bib:hsiao_rss_2010}~\cite{bib:platt_csail_2011};

\item[Large action branching factor] increasing the complexity of the robot manipulator leads to a higher dimensional action space. The planning problem with a single finger is to search for a sequence of pushes that will move the object from a start to a goal configuration. The action space is thus already a high dimensional continuous space. We have solved this problem randomly sampling possible linear finger trajectories which ensure that the end effector of the manipulator will collide with the object. In grasping, however, actions are no longer defined as straight line trajectories, but rather as more complex rotational and transitional movements of the wrist and fingers which now compose the end effector of the manipulator;

\item[Large observation branching factor] observations we can take into consideration while grasping assume the form of torque/force sensors, vision, proprioception, tactile sensors and so on. All of these observations provide a continuous or near-continuous representation of the current configuration of the environment. The planning problem requires the ability to predict observations in future stages in order to define the best policy. A large observation branching factor thus will reduce performances and quality of the solution.
\end{description}

Our approach is to study and possibly extend a new approach to artificial intelligence (A.I.) which has emerged in the last years in response to the necessity of building intelligent controllers for agents operating in unstructured stochastic environments (see i.e.~\cite{bib:hsiao_rss_2010}~\cite{bib:platt_csail_2011}~\cite{bib:hauser_wafr_2010}).  Such agents require the ability to learn by interaction with its environment an optimal action-selection behaviour. The main issue is that real-world problems are usually dynamic and unpredictable. Thus, the agent needs to update constantly its current image of the world using its sensors, which provide only a noisy description of the surrounding environment. Although there are different schools of thinking, with their own set of techniques, a brand new direction which unifies many A.I. researches is to formalise such agent/environment interactions as \emph{embedded systems} with stochastic dynamics.

In the following subsections of this introduction, we will briefly discuss the difficulties inherent in designing controllers for autonomous robotic manipulators and, in particular, we shall argue why adaptive controllers are attractive in such environments. We will also introduce the working scenario as a list of hardware components that will be used to develop this thesis project.

\subsection{Deriving Controllers for Embedded Systems}

The systems as mentioned above are referred to in the literature as embedded systems. Many researchers are unified by the belief that such systems in which the interaction between the agent and its environment is a predominant factor should be designed in terms of dynamical systems rather than as composed of two independent objects~\cite{bib:jeremy}. In other words, the agent-environment interaction can be viewed as a coupled dynamic system in which the outputs of the agent influence the environment as inputs and the outputs of the environment affect the agent as inputs. In fact, the agent's outputs are its planned actions which alter the surrounding environment and the environment's outputs are perceived by the agent's sensors. As already mentioned, it may be the case that the agent fails to build a complete description of the state of the environment because of either noisy perceptions or a non-static environment which changes asynchronously with respect to the agent's perception timing. Finally, it is important to note that in embedded systems the future sequence of states of the environment is influenced by the agent's outputs at each stage.

From the perspective of a designer of intelligent agents for embedded systems, the aforementioned characteristics have several implications. First, an incomplete description of the state of the environment forces the agent to act in partial ignorance. Second, when the environment is not passive and its evolution is not entirely influenced by the agent's inputs, the agent has an upper bound on the time it takes to decide what to do next. 
Furthermore, since earlier decisions of the agent may significantly influence its future internal states, the embedded intelligent agent should be designed in terms of an adaptive agent. More precisely, we should take into consideration an adaptive agent's mapping function from the inputs to outputs. Such a mapping is usually called a \emph{policy} in literature.

A branch of control theory addresses the set of problems concerning non-static (or dynamic) environments which keep changing through time in a non-deterministic way, supplying methods for building adaptive control policies. These methods define \emph{closed-loop} control policies which operate in such models specifying the evolution of the system according to the state of the environment and its underlying task. Closed-loop policies are therefore suitable for controlling stochastic systems. Phenomena of the system such as internal state transitions and action outcomes can be modelled as stochastic functions according to a probability density function (pdf) which is usually specific to the domain of application. Furthermore, most of the problems can be expressed assuming that the evolution of the system depends only on the previous internal state and independently of the previous history. This property is known as the \emph{Markovian Property} and it defines the basic assumption of a rich framework, termed \emph{Markov Processes (MP)}, for embedded systems with stochastic outcomes.

An adaptive agent may learn how to modify its own policy according to either specific domain knowledge or its experience. A way to create a learning agent is to include in the agent's inputs also some sort of feedback other than an observation of the state of the environment. Feedback may be structured as signals describing correct behaviours and penalising wrong choices. Such a pattern is known as supervised learning and assumes an entity such as a teacher who knows exactly the optimal behaviour. Alternatively, the feedback may evaluate the agent's policy by some \emph{Index of Performance (IoP)}. In this case, no teachers are necessary and the evaluation is based on some objective criteria which describe merely how good (or bad) the agent's policy is. The latter approach is useful for a wider range of problems where the optimal behaviour is incognito and it is known as \emph{Reinforcement Learning (RL)}.

In short, reinforcement learning defines a trial and error approach in which the agent learns how to interact with the surrounding environment. The learning is task-dependent. Correct behaviours of the agent receive positive feedback termed \emph{reinforcement}. Generally, this feedback is a sequence of scalar values that express the ``goodness'' of a particular behaviour. The mapping between states of the environment and agent's actions to reinforcement values is known as the \emph{reinforcement} or \emph{reward function}.

\subsection{Motion Planning in Unstructured Environments}

Littman in~\cite{bib:littman} describes \newdef{sequential decision making} using the following examples:
\begin{quote}
A frog jumps around a barrier to get to a delicious mealworm. A commuter tries an unexplored route to work and ends up having to stop and ask for directions. A major airline lowers prices for its overseas flights to try to increase demand. A pizza-delivery company begins a month-long advertising blitz. These are examples of sequential decision making.~\cite{bib:littman}
\end{quote}
In other words, a frog which plans its behaviour for the next time step in order to achieve its task (eating a delicious mealworm) is a problem that can be formally modelled as sequential decision making. The actions selected by the frog are defined as jumps and its behaviour or \emph{policy} might be described as the need to jump around the barrier and reach the prey with as few jumps as possible. At any time we expect that our frog will select the best jump in order to go round the obstacle, following as closely as possible the optimal ``trajectory'' (i.e. the one which requires the fewest number of jumps). The frog will then  learn its optimal behaviour by experience, jumping and (succeeding in) eating mealworms; in fact when its policy chooses actions which lead to sub-optimal trajectories it might not capture its prey, discouraging the frog from following  such behaviour again. %Such paradigm is termed \newdef{Reinforcement Learning}.

There is another aspect of the problem that the frog should take into consideration, that is how to execute the selected action. In fact the jump should involve a reliable trajectory which can be actually executed by the frog. In other words, the frog can execute such a jump with the intended force and direction; and the whole trajectory should be in a free space where no obstacle can jeopardise the outcome. In robotics, that is a well known problem that can be solved by a full branch of techniques and algorithms termed in literature as \newdef{motion planning}.

Recently, progress has been made on the problem of planning motion for robots with many degrees of freedom through complex environments~\cite{stuber2020,bib:zito_2019,bib:zito_2013,bib:zito_w2013,bib:zito_2012,bib:rosales_2018,bib:zito_ecvp_2012,bib:zito_w2012,bib:stuberICRA2018,bib:barskyWIROS2018,heiwoltICORR2019,bib:denoun2019,bib:zitoWRRS2019a,bib:zitoWRRS2019b,bib:zitoWRRS2019c}. However, such robots are only used in carefully structured domains, where poses and shapes of objects within the working space of the robot are exactly known \emph{a priori}. Usually, the environment is assumed to be static which guarantees that its configuration does not change according to external forces.  In this case robots are faster and produce a more accurate job than any human. In real-world problems however, such assumptions must be released and the agent needs to infer configurations and dynamics of the surrounding environment  using its sensors and previously gathered experience. Sensors provide noisy signals and required calibrations which are difficult to achieve.
Anyway, we mainly aim to deal  only with the uncertainty on the pose of the object to manipulate. Therefore we will assume our working scenario to be exactly known, except for the initial pose of the object which is inferred using vision sensors. The scenario is also assumed to be static, which means that it is not subject to external forces and it can not change its configuration unless there is an explicit action from the agent. Nevertheless, the dynamics of the system provide further uncertainty to the problem because they are non-linear and hard to approximate.  
We are also interested to model our system with stochastic dynamics in order to deal with unstructured world\footnote{Probabilistic approaches are not the only ones reliable. E.g. closed-loop policies might be only based on sensor feedback. However, probabilities express our confidence over our current knowledge of the world and may lead to more efficient solutions.}. 

To motivate the methods we aim to use in this thesis, we make the following observation: the fundamental problem of planning in scenarios where the agent needs to interact with its environment by making contacts with (fully or partially) unknown objects  is that the configurations of the robot and objects in the world are not exactly known at the time of execution. We therefore plan not only how to change the world with our actions but to change also what we know about the world. In other words, the planner or controller should tell us how confident we are with the representation of relevant aspects of the world, for example whether or not the agent achieved a stable grasp. Consequently our goal is designing a planner which leads to a subset of states where this confidence is maximised in order to optimally solve the intended task. The problem requires the ability to balance confidence maximisation and use of the previously acquired knowledge. Formally, this is termed the \newdef{exploration-exploitation dilemma}.

\subsection{The Belief Space}

In order to design such a planner, we aim to develop an \emph{adaptive} agent which learns how to behave in its environment to maximise the expectation of succeed. We aim to express what we know about the uncertain aspects of the world as a \emph{belief state}, which is the probability density distribution (pdf)  over the possible set of world states, and then select actions according to the current belief state. In other words, we will use a stochastic approach to model uncertainties. The agent will then have to find the optimal action or \emph{control input} which maximises our expectation of completing the task. Therefore we will attempt to solve the problem of grasping and manipulation using algorithms and techniques from the \newdef{stochastic control theory}. 

\begin{figure}[tb]
  \centering
  \includegraphics[width=4in]{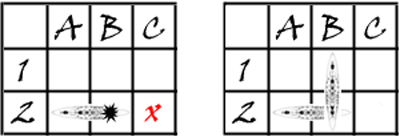}
  \caption{Tiny example of representing a Battleship instance with a belief state. The left hand picture shows the rea position of the 2-block ship (A2 and B2). After observing a ``miss'' in C2 and a ``hit'' in B2 we can express our belief state as probability distribution over the remain possible displacements of the ship, as shown on the right hand of the picture.}\label{fig:battleship}
\end{figure}

As a concrete example of how such a model might look like, we consider a simple instance of the game of Battleship. This is a wildly used example in A.I. community (see e.g.~\cite{bib:hsiao_thesis}). In our tiny example, we play a 2 by 3 scheme, one-side Battleship. The goal is to destroy the only 2-block ship in the scheme. The left side of Figure~\ref{fig:battleship} shows our example on which the ship is placed in A2 and B2. In the beginning, the agent has no clue where the ship could be. Therefore, the agent might act guessing C3 as first ``shot'' and then observe a ``miss''.  Successively it might guess B2 and observe a ``hit''.  A feasible way to express our knowledge of the problem after the second observation is as a probability distribution over possible displacements of the ship, which is our belief state. The right side of Figure~\ref{fig:battleship} shows our belief state after these two actions. In fact after observing a ``hit'' only two displacements are possible for the ship, each with probability of 1/2. We now expect that the player selects the next action according to the current belief state, which will be B1 or A2 with same probability.

As also shown in~\cite{bib:hsiao_thesis}, we might want to use the same representation to describe a different problem. We can easily imagine that the ship in our Battleship example might be a sitting object on a table. The goal now is to achieve a stable grasp of the object. In this case, the agent perceives the world by noisy sensors  (e.g. vision system) and the initial uncertainty about the pose of the object might be high enough to make traditional open-loop plans (even extended with simple feedback) not reliable. We therefore aim to reduce the initial uncertainty applying \emph{information gathering} actions which lead to maximise the likelihood of success. However,  differently from the Battleship example, our object  is unlikely to be only in few discrete locations on the table and touching the object in order to estimate better its pose can cause it to move.

\begin{figure}[ht]
  \centering
  \includegraphics[width=4in]{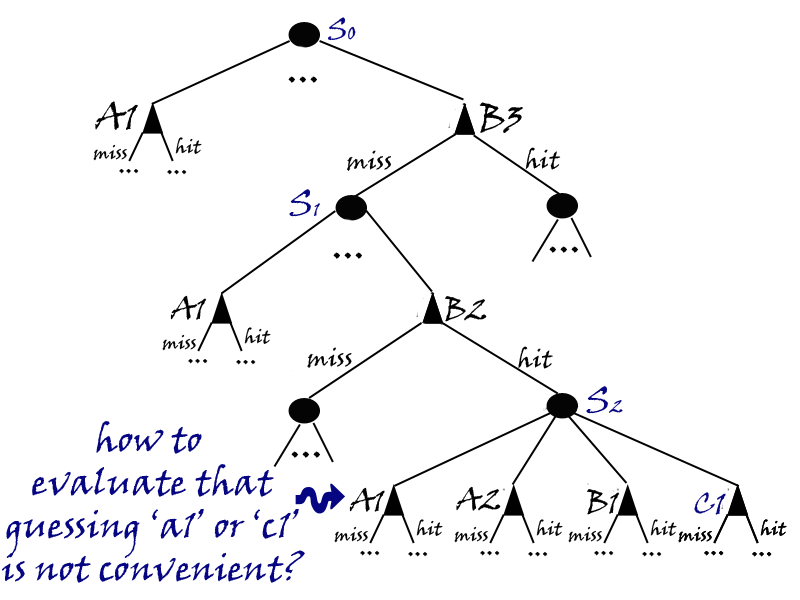}
  \caption{Search tree for the Battleship example. Circles represent the configurations or states of the game. The root hence represents the initial state ($S_{0}$). Triangles represent actions, that is ``guess for the player''. For each action only two possible observations are available: ``hit'' or ``miss''. The picture shows only three levels of the tree according to the evolution of the game ($S_{0}$,B3,miss,$S_{1}$,B2,hit,$S_{2}$); In order to extend the state $S_{2}$ there are only four possible actions applicable. The main question is how the planner can evaluate that guessing A1 or C1 is not convenient at this level of the game.}\label{fig:battleship_tree}
\end{figure}

\subsection{Modeling Uncertainty with Markov Processes}

\newdef{Partially Observable Markov Decision Processes (POMDPs)}~\cite{bib:littman}~\cite{bib:pomdp}, described in detail in Section~\ref{sec:pomdp}, are a popular framework to solve decision making problems with uncertainty. In order to solve exactly a POMDP is necessary to compute \emph{off line} a policy which describes the optimal agent’s behaviour for any possible state of the problem.  
For example, it is possible to reason in terms of POMDP formulation to model the Battleship problem previously proposed. Figure~\ref{fig:battleship_tree} shows how the search tree for the Battleship example looks like. The tree is trunked at the third level which defines the situation after two moves. The state labeled as $S_{2}$ identifies our current situation after  observing a ``hit'' in B2, where four actions are available but only two of them are feasible. The off line POMDP solver would extend the root with all possible evolutions of the game to return a policy which identifies the best action according to the previous history of the game. In literature a precise evolution of the problem as the one shown in Figure~\ref{fig:battleship_tree} with the sequence of states $<S_{0},S_{1},S_{2}>$ is often termed as \newdef{trajectory in the solution space}.

Many real problems, therefore, are not suitable for this kind of approach because they present large or continuous state, action and observation spaces. Although some approximations techniques have been developed (see i.e.~\cite{bib:spaan_perseus}),  the state-of-the-art of planning in high-dimensional continuous spaces would require us to draw only a random subset of possible trajectories in the solution space and follow the most promising one to reach a possible final configuration. Such a method is known as \newdef{Monte-Carlo Search Tree (MCST)} and has been applied successfully also to problems model as POMDPs~\cite{bib:silver}. In this work, the policy is computed \emph{on line} alternating planning phase to execution phase in order to alleviate computational complexity focusing only on local policies. In fact, online approaches are forward search rooted in the current belief state; such approach does not guarantee optimality but reduces the branching factor of the search because it only needs to consider the reachable belief states from the current belief state.
Although different optimization methods have been proposed and good performance has been reached for many difficult problems, as in $9\times9$ Go~\cite{bib:utc}, these methods are based on look-ahead strategies which involve prediction on the most likely system evolutions. Consequently, it is hard to model long-term trajectories in systems with strong uncertainty related to the actions' outcome. 

\subsection{Related Work}

Platt et al. in~\cite{bib:platt_csail_2011} propose a formalization of the problem of grasping in unstructured environments as \newdef{simultaneous localisation and grasping (SLAG)}. SLAG can be viewed as a belief space control problem. In other world, this formalization allows them to figure out a sequence of motor control actions for grasping while bounding the probability of failure. This approach has been applied to a real scenario on which the robot needs to localise and grasp a box. Two boxes of unknown dimension are presented to the robot. This situation is challenging because the displacement of the two boxes might make harder the problem of identifying the exact pose of the boxes themselves. In their example, the robot is equipped with two paddles and it has a pre-programmed ``lift'' function. The robot localises objects using a laser sensor mounted on its left wrist. The boxes, however, are assumed to be placed at a known height and thus the robot has uncertainty only in one dimension. 

Hsiao et alii in~\cite{bib:hsiao_rss_2010} show a tactile-driven exploration of the environment to maximise the expected confidence relative to the object pose before attempt a grasp. In this work, the set of actions is parameterised with the current belief state over the object’s pose. A fixed number of actions for gathering information and grasp are pre-computed offline. The planner solves the POMDP on line selecting gathering information actions which reduce the uncertainty about the object’s pose until a threshold of confidence has been reached, and then the pre-computed grasp action is executed. If the belief state after all the gathering information actions correctly approximate the real object’s pose, the grasp has high probability to succeed.

Hauser in~\cite{bib:hauser_icml_2011} and~\cite{bib:hauser_wafr_2010} present a sample-based replanning strategy for driving partially observable, high-dimensional robotic systems to a desired configuration. The planning algorithm uses forward simulation of randomly sampled open-loop controls in order to build a search tree in the belief space. At any time the search tree is rooted on the current belief state. Then the algorithm select the action from the root which leads to the best evaluated node in the tree. The procedure is repeated until the goal region is reached. Hauser performed experiments on two scenarios: a 2D pursuit scenario with 4D state space; and a localisation scenario in a known $d$-dimensional environment, on which the convergence of the algorithm has been demonstrated up to 7 dimensions.
%Petrovskaya at alii in~\cite{bib:petrovskaya_icra_2006}~\cite{bib:petrovskaya_jcrai_2007} propose a Bayesian approach to estimate position and orientation of objects from tactile sensors. With a top-down fashion technique, they performs a search using a series of refinements in order to gradually scale the precision from low to high with proof of concept in application to manipulating a box and grasping a door handle. 

\subsection{Our Approach}

We are interested to solve the grasping problem using online strategies as Hsiao et al. do, but without the limitation of a fixed set of actions. We will rather sample random actions at planning time, and we will evaluate our belief state according to the executed simulations. 

In this thesis, we address to solve the problem of planning with uncertainty in stochastic environments using reinforcement learning techniques. Section \ref{sec:solution} discusses in more detail our approach. 

\subsection{Working scenarios}\label{sec:workingscenario}

The grasping scenario we have been working on is composed as follows:
\begin{enumerate}
\item Kuka arm KR 5 sixx R850 6DOF
\item DLR-Hand II prototype of five-fingered hand;
\item a camera for vision capture;
\item torque sensors on all finger joints;
\item force torque sensors at wrist;
\item a known object to be manipulated.
\end{enumerate}

The pushing scenario we have been working on is composed as follows:
\begin{enumerate}
\item 5-axis Katana 6M180 arm equipped with a spherical probe as end effector;
\item a camera for vision capture;
\item 6-axis force torque sensor at the base of the end effector;
\item a known object to be manipulated (polyflap).
\end{enumerate}

\subsection{Goals to be achieved}
Such a division should show how the grasping and manipulation problem could be split in order to define the problem space of my thesis. The subtasks are as follows:
\begin{enumerate}
\item attempt a stable grasp;
\item move objects from one configuration to another on a table;
\item move objects, held and supported entirely by the hand, from one configuration to another with respect the frame of the hand (in-hand manipulation).
\end{enumerate}

\subsection{Structure of this report}

This report is structured as follows: 
\begin{itemize}
\item section~\ref{sec:problem} describes the problem domain of planning with uncertainty for autonomous robotic manipulators;
%how the grasping and manipulation space is defined;
\item section~\ref{sec:solution} first introduces models for path planning in high-dimension continuous spaces and models for planning with uncertainty. It then presents how to use specific domain knowledge to obtain better performance and, finally, it shows some hybrid models we have taken inspiration from for our proposed solution;
\item section~\ref{sec:adaptivecontrol} first introduces some technical detail specific of our domain and then shows formally our approach. 
\item section~\ref{sec:evaluation} describes how we will evaluate the performance of our algorithms. 
\end{itemize}

%------------------------------------------------------------------------------------------------------------------------------------------------------------------------------------------------
%	PROBLEM SPACE
%------------------------------------------------------------------------------------------------------------------------------------------------------------------------------------------------
\section{Problem domain}\label{sec:problem}

In this section, we present the problem domain relative to the problem of grasping and manipulating objects using a robotic hand.

We are concerned with solving problems involving an intelligence agent acting autonomously in a dynamic environment. More properly, we design the agent-environment interaction as an embedded system rather than two single, independent elements. Our embedded system will assume the following characteristics:
\begin{enumerate}
\item quasi-static conditions;
\item deterministic knowledge of the agent's end effector position in the working space;
\item uncertainty related to the pose and/or shape of the object;
\item uncertainty related to the outcome of the agent's actions.
\end{enumerate}
In many robotic operations, manipulations are executed at a safe speed and one can assume that the evolution at the next time stage of two interacting bodies can be entirely described by only the current state of the system, ignoring all the previous internal states of the system. This characteristic is known as a \emph{quasi-static condition}~\cite{bib:kopicki_2010}. 
We can express the assumptions of the points 2-4 as the agent's position in the global frame is reasonably well known, but that there is some uncertainty about the relative pose and/or shape of the object to be manipulated. Additionally, we assume that there are force sensors on the robot that can reasonably detect when it makes or loses contact. Hence, we assume that a reasonably accurate model of the task dynamics and sensors is known and the principal uncertainty is in the configuration of the robot and the state of the objects in the world. 
The last point expresses our uncertainty on how the agent's output influence the surrounding environment. That directly affects our ability to predict system evolutions in according to the internal state and the action executed.

%---------------------------------------------------------------------------------------------------------------------
%	GRASPING
%---------------------------------------------------------------------------------------------------------------------
\subsection{Grasping}

Grasping an object involves (1) determining a set of contact points for the fingers on the surface of the object in order to achieve a stable grasp; (2) planning a trajectory of the arm to bring the hand in a pre-grasp position where it is possible to conclude the grasp by closing the fingers; and (3) a finger closing strategy which copes with the kinematic constraints of the hand. 

The problem of grasping objects is an hard challenge for many reasons. First of all, even for known objects and a given grasp it is hard to define in an analytic way the exact configuration of the hand for which only the expected surfaces collide in the predicted way. It is obviously more difficult when objects are unknown or partially visible: for example, how to predict whether or not an unknown mug has an handle on the opposite side with respect to the camera?  
Additionally, planning for grasping requires to model a sequence of non-trivial physic interactions between fingers/palm of the robot hand and the object as, for example, precocious contacts which modify in somehow the world configuration making inadequate the pre-computed planning. 

%---------------------------------------------------------------------------------------------------------------------
% 	MANIPULATION
%---------------------------------------------------------------------------------------------------------------------
\subsection{Manipulation}

The manipulation of an object with a robotic multi-fingered hand can be split into three different categories: (1) single finger manipulation (or pushing); (2) multi-finger manipulation of an object on a stable surface; and (3) in-hand manipulation of an object held and supported entirely by the hand. 
The single finger manipulation concerns simple pushing an object from a robotic arm using a simple end-effector which generates only a single point of contact with the object. In this case, the aim is to plan a sequence of actions (or pushes) to move the object from an initial pose to a desired one.
The multi-finger manipulation on a stable surface can be seen as an extension of the case described above where multiple contacts are allowed. However, it is not a trivial extension because the result of a combination of actions applied at the same time is generally different from what we would obtain applying the same sequence of actions separately. However in-hand manipulation involves manipulating an object within one hand. The fingers and the thumb are used to best position the object for the required activity, for example, picking up a pen and moving it into the position with your fingers for writing. In other words, in-hand manipulation is the result of a combination of forces (or pushes) applied by several fingers to a target object in order to achieve a desired object pose.

All these manipulations necessitate the ability to predict how objects behave under finger manipulation and, furthermore, the ability to plan a sequence of actions to move the object from the initial pose to the desired one. At the time of writing this report, we are developing a simple path planner to cope with a physical environment and to achieve a desired pose of a given object using single finger pushing operations.

%------------------------------------------------------------------------------------------------------------------------------------------------------------------------------------------------
%	SOLUTION SPACE
%------------------------------------------------------------------------------------------------------------------------------------------------------------------------------------------------
\section{Solution space}\label{sec:solution}

In this section, we present a possible solution space for planning in which the robot has to interact through contact with the surrounding world.

In robotics, as described in~\cite{bib:lavalle_2006}, motion planning was originally concerned with problems such as how to move a piano from one room to another in a house without hitting anything. However, the field has grown to include complications such as uncertainties, multiple bodies and dynamics. In artificial intelligence, planning originally meant a search for sequence of logical operators or actions that transform an initial world state into a desired goal state. Presently, planning extends beyond this to include many decision-theoretic ideas such as Markov decision processes and imperfect state information.

Up until now, we have considered two approaches to the problem, namely: (1) techniques for planning in continuous state and action spaces and (2) planning in uncertain worlds.
In the following subsections, we will briefly introduce some models for both the approaches, and we will introduce an innovative way to combine this techniques into an hybrid model.

\begin{figure}[t]
  \centering
  \includegraphics[width=3in]{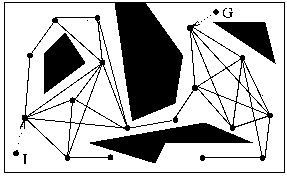}
  \caption{An illustration of the probabilistic network [URL:http://www.kavrakilab.org/robotics/prm.html].}\label{fig:prm}
\end{figure}
%---------------------------------------------------------------------------------------------------------------------
% 	PLANNING IN CONTINUOUS SPACE
%---------------------------------------------------------------------------------------------------------------------
\subsection{Planning in continuous spaces}  

%---------------------------------------------------------------------------------------------------------------------
The {\bfseries Probabilistic RoadMap (PRM)}~\cite{bib:kavraki_1996}~\cite{bib:geraerts_2002} planner is a motion planning algorithm in robotics, which solves the problem of determining a path between a starting configuration of the robot and a goal configuration whilst avoiding collisions.

The basic idea behind PRMs is to take random samples from the configuration space of the robot, test them for whether they are in free space, and use a local planner to attempt to connect these configurations to other nearby configurations. The starting and goal configurations are added in, and a graph search algorithm is applied to the resulting graph to determine a path between the starting and goal configurations.

The probabilistic roadmap planner consists of two phases: a construction and a query phase. In the construction phase, a roadmap (graph) is built, approximating the motions that can be made in the environment. First, a random configuration is created which is then connected to some neighbours, typically either the k nearest neighbours or all neighbours less than some predetermined distance. Configurations and connections are added to the graph until the roadmap is dense enough. In the query phase, the start and goal configurations are connected to the graph, and a path is obtained using Dijkstra's shortest path query.

The main disadvantage of the PRM is that this model seems impractical for many problems with non-holonomic\footnote{A system is non-holonomic if the controllable degrees of freedom are fewer than the total degrees of freedom} constraints. 

%---------------------------------------------------------------------------------------------------------------------
A {\bfseries Rapidly-exploring Random Tree (RRT)}~\cite{bib:lavalle_2001}~\cite{bib:lavalle_2006} is a data structure and algorithm that is designed for efficiently searching non-convex high-dimensional spaces. RRTs are constructed incrementally in a way that quickly reduces the expected distance to a randomly-chosen point in the tree. RRTs are particularly suited for path planning problems that involve obstacles and differential constraints (non-holonomic or kino-dynamic). RRTs can be considered as a technique for generating open-loop trajectories for nonlinear systems with state constraints. An RRT can be intuitively considered as a Monte-Carlo way of biasing search into the largest Voronoi regions.
\begin{figure}[t]
  \centering
  \includegraphics[width=2.0in]{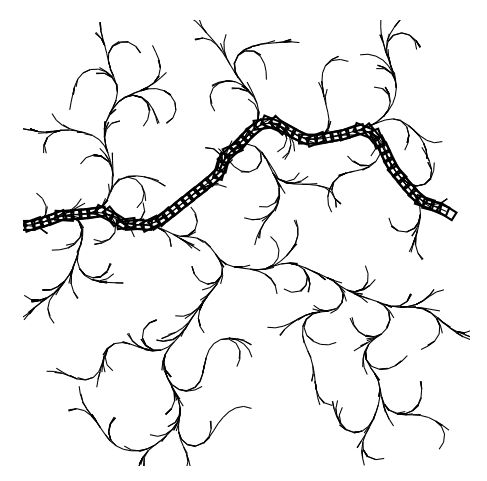}
  \caption{A 2D projection of a 5D RRT for a kinodynamic car~\cite{bib:lavalle_2006}.}\label{fig:rrt}
\end{figure}

Unlike PRMs,  RRTs cope well with non-holonomic constraints. However, both techniques do not model the uncertainty of the environment in a primitive way. Recently, some techniques have been developed to make RRTs work in uncertain worlds. These techniques are discussed in the section~\ref{sec:prrt}. Additionally, PRMs is not designed to explore a continuous space seeking for a possible solution. PRMs are used to grow multiple distributed trees and can be used for multiple queries to find path between two given points. The underlying assumption is that the goal position is not part of the tree and should be known as well as the initial position. Whilst, RRTs is a proper exploring algorithm which is able to explore high-dimensional spaces seeking for the point which satisfies all the problem constraints.

%---------------------------------------------------------------------------------------------------------------------
% 	PLANNING IN UNCERTAIN WORLDS
%---------------------------------------------------------------------------------------------------------------------
\subsection{Planning in uncertain worlds}\label{sec:pomdp}

\begin{figure}[t]
  \centering
  \includegraphics[width=2.0in]{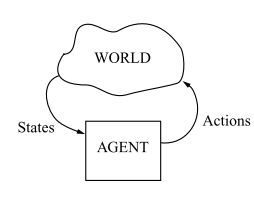}
  \caption{An MDP models the synchronous interaction between agent and world~\cite{bib:littman}.}\label{fig:mdp}
\end{figure}

%---------------------------------------------------------------------------------------------------------------------
An {\bfseries Markov Decision Problem (MDP)}~\cite{bib:littman}~\cite{bib:pomdp} is a model of an agent interacting synchronously with a world. As shown in Figure~\ref{fig:mdp}, the agent takes as input the state of the world and generates as output actions which themselves affect the state of the world. In the MDP framework, it is assumed that, although there may be a great deal of uncertainty about the effects of an agent's actions, there is never any uncertainty about the agent's current state, it has complete and perfect perceptual abilities.

A discrete stochastic process is defined by a set of random variables $\{X_{t}, t\in\mathcal{T}\}$, where $\mathcal{T}=\{0,1,2,\dots\}$ is the set of possible times and $X_{t}$ denotes the ouotcome at the $t^{th}$ stage or time step. The domain of $X_{t}$ is the set of all the possible outcomes denotes $\mathcal{S}=\{s_{1},s_{2},\dots,s_{|\mathcal{S}|}\}$
A MDP can be described as a tuple $\langle\mathcal{S},\mathcal{U},\tau,\rho\rangle$, where
\begin{itemize}
\item $\mathcal{S}$ is a finite set of states of the world;
\item $\mathcal{U}$ is a finite set of actions;
\item $\tau\colon\mathcal{S}\times\mathcal{U}\rightarrow\Pi(\mathcal{S})$ is the state transition function, giving for each world state and agent action, a probability distribution over the world's states;
\item $\rho\colon\mathcal{S}\times\mathcal{U}\rightarrow\Re$ is the reward function, giving the expected immediate reward gained by the agent for taking each action in each state.
\end{itemize}

In accordance with the Markov property, the next state and the expected reward depend only on the previous state and the action taken; even if it conditions the additional previous states, the transition probabilities and the expected rewards would remain the same. 
%---------------------------------------------------------------------------------------------------------------------
While the MDP addresses the problem of choosing optimal actions in \emph{completely observable stochastic domains}~\cite{bib:em}, we need a model more suited with uncertainty. For example, in the grasping problem, once the robot has perceived the object to grasp (using vision and the 3D geometric features) and computed a path to reach the pre-grasp position, the robot should define and plan a sequence of actions to close the fingers in a proper way to maximize the likelihood of successful grasp. For doing this, the robot needs a formal model of sequential decision making which describes the environment with which it interacts and the behavior it exhibits. Furthermore, this model should be able to cope with uncertainty because the world around the robot has perceived using noisy sensors. In practice, we are considering the problem of choosing optimal actions in \emph{partially observable stochastic domains}. Problems like the one describe above can be modeled as {\bfseries Partially Observable Markov Decision Processes (POMDPs)} ~\cite{bib:littman}~\cite{bib:pomdp}.
In POMDPs, the system interacts with a stochastic environment whose state is only
partially observable. Actions change the state of the environment and lead to numerical
penalties/rewards, which may be observed with an unknown temporal delay. The model
aims to devise a policy for action selection that maximizes the reward.

Formally, a POMDP can be formalized as a tuple $\langle\mathcal{S},\mathcal{U},\mathcal{Y},\tau,\rho,\psi\rangle$, where $\mathcal{S},\mathcal{U},\tau$ and $\rho$ keep the same definition aforementioned. In addition:
\begin{itemize}
\item $\mathcal{Y}$ is a finite set of observations;
\item $\psi\colon\mathcal{S}\times\mathcal{U}\rightarrow\Pi(\mathcal{Y})$ is the observation function, giving for each world state and agent action, a probability distribution over the set of observations;
\end{itemize}

Obviously, the POMDP framework embraces a large range of practical problems. However, solving a POMDP is often intractable except for small problem due to their computational complexity: finite-horizon POMDPs are PSPACE-complete~\cite{bib:papadimitriou} and infinite-horizon POMDPs are undecidable~\cite{bib:madani}. 

\subsubsection{POMDP Framework}\label{sec:pomdpframework}

The key aspect in POMDPs is the assumption that the agent has no direct access to the states of the environment. It has to infer its knowledge from some observation that gives incomplete information about the current state. In accordance with~\cite{bib:pineau}, a complete history of the system at time t is define as:
\begin{equation}\label{eq:history}
h_{t}=\{u_{0},y_{1},\dots,y_{t-1},u_{t-1},y_{t}\}
\end{equation}
This explicit representation of the past is typically memory expensive. It is possible to summarize all the relevant information in a probability distribution over the state space $\mathcal{S}$. In literature, the probability distribution over states (as well as over actions or observations) is referred as a belief state and the entire probability space (the set of all possible probability distributions) as the \emph{belief space}. A belief state at time $t$ is defined as the posterior distribution of being in each state, given the complete history:
\begin{equation}\label{eq:belief}
b_{t}(x)=\Pr(x_{t}=s|h_{t},b_{0})\quad\forall s\in\mathcal{S}
\end{equation}
The belief $b_{t}$ is a sufficient statistic for the history $h_{t}$~\cite{bib:smallwood}. Therefore, the agent is able to choose the current action in according to its current belief state and the initial belief, $b_{0}$. At any time $t$, the belief state $b_{t}$ can be computed following the \emph{Bayesian filtering} from the previous belief state $b_{t-1}$, the previous action $u_{t-1}$ and the current observation $y_{t}$, as follows:
\begin{equation}\label{eq:filtering}
b_{t}(x_{t}=s')=\Pr(y_{t}|x_{t}=s')\sum_{s\in\mathcal{S}}{\Pr(x_{t}=s'|u_{t-1},x_{t-1}=s)b_{t-1}(x_{t-1}=s)}
\end{equation}
Once we define a way to compute the current agent's belief state, the important question is how to use this information for choosing an action at any time $t$. This action is determine by the policy $\pi$ which specifies the probability to use any action in any given belief state. In other words, the policy defines the agent's strategy for all the possible situations it may encounter. An optimal strategy should maximise the expected sum of discounted rewards over the time $T$, as follows: 
\begin{equation}\label{eq:policy}
\pi_{T}^{*}=\arg\max_{\pi\in{\Pi}}E[\sum_{t=0}^{T}\gamma^{t}\sum_{x\in{\mathcal{X}}}b_{t}(x)\sum_{u\in{\mathcal{U}}}\rho(x,u)\pi(b_{t}, u)|b_{0}]
\end{equation}
Where $\gamma\in[0,1)$ is the discount factor and $\pi(b_{t}, u)$ is the probability that action $u$ is performed in belief $b_{t}$ according to the policy $\pi$. In a similar way, we also compute the reward obtained following the policy $\pi$:
\begin{equation}\label{eq:reward}
V^{\pi}(b)=\sum_{u\in\mathcal{U}}\pi(b,u)[R(b,u)+\gamma\sum_{y\in{Y}}\Pr(y|b,u)V^{*}(\tau_{y}(b,u))]
\end{equation}
where $R(b,u)$ is the immediate expected reward.
The optimal policy $\pi^{*}$ defined in the equation~\ref{eq:policy} represent the strategy that maximise the equation~\ref{eq:reward}, formally we write as: 
\begin{equation}\label{eq:opvalue}
V^{*}(b)=\max_{u\in\mathcal{U}}{[R(b,u)+\gamma\sum_{y\in{Y}}\Pr(y|b,u)V^{*}(\tau_{y}(b,u))}]
\end{equation}
Another useful quantity is the Q-value which defines the value of \emph{u} by assuming that the optimal policy is followed at every step afterwards.
\begin{equation}\label{eq:qvalue}
Q^{*}(b,u)=R(b,u)+\gamma\sum_{y\in{Y}}\Pr(y|b,u)V^{*}(\tau_{y}(b,u))
\end{equation}

\subsubsection{Acting Optimally in Discrete Stochastic Environments}\label{sec:actingoptimally}

The value in equation~\ref{eq:reward} is also known as \emph{stochastic Bellman equation} or \emph{dynamic programming equation}. The idea is to break down the long-period planning into simpler stage-wise steps. Moreover, for the \emph{Bellman's principle of optimality} every segment of an optimal path is itself optimal~\cite{bib:bellman}.
\begin{figure}[t]
  \centering
  \includegraphics[width=4in]{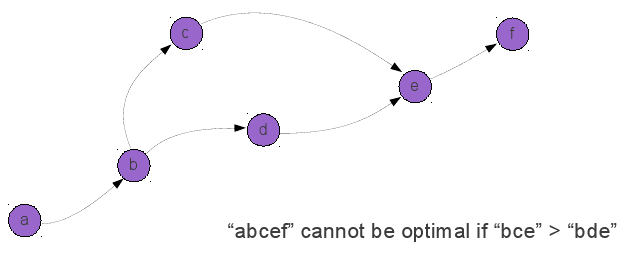}
  \caption{Optimal sub-structure~\cite{bib:tutorialcandido}}\label{fig:bellman}
\end{figure}
A key result presented in~\cite{bib:smallwood} shows that the optimal value function for a finite-horizon POMDP can be represent by hyperplanes, and therefore is a piecewise linear and convex function. The hyperplanes are also known as $\alpha$-vectors. Unfortunately, the number of $\alpha$-vectors the solving algorithm has to evaluate grows exponentially in the number of the observations at each iteration.  The complexity for compute the exact solution of iteration $t$ is $O(|\mathcal{S}|^{2}|\mathcal{U}||\mathcal{Y}||\Gamma_{t-1}|^{|\mathcal{Y}|})$ where $\Gamma_{t-1}$ represents the set of $\alpha$-vectors at time $t-1$~\cite{bib:pineau}. 

A new approach to Artificial Intelligence has emerged in the last decades attempting to improve the applicability of POMDP approaches to larger problems by developing approximate offline techniques. Here, we propose a quick summarising (for further details see~\cite{bib:pineau}):
\begin{itemize}
    \item Point-Based Value Iteration (PBVI)
    \begin{itemize}
        \item Maintains only a set of belief states
        \item Only considers constraints that maximize the VI for at least one example 
        \item Define a lower bound
    \end{itemize}
    \item MDP
    \begin{itemize}
        \item Does not consider the uncertainty of the states
        \item Defines an upper bound 
    \end{itemize}
    \item QMDP
    \begin{itemize}
        \item Defines a single $\alpha$-vector for each action
        \item The uncertainty of the states disappears after a single step
        \item Defines an upper bound 
    \end{itemize}
    \item Fast Informed Bound (FIB)
    \begin{itemize}
        \item Defines a single $\alpha$-vector for each action
        \item Belief state is updated taking into account (at some degree) the partially observability of the environment
        \item Defines a tighter upper bound than QMDP 
    \end{itemize}
\end{itemize}
\begin{figure}[ht]
  \centering
  \includegraphics[width=4in]{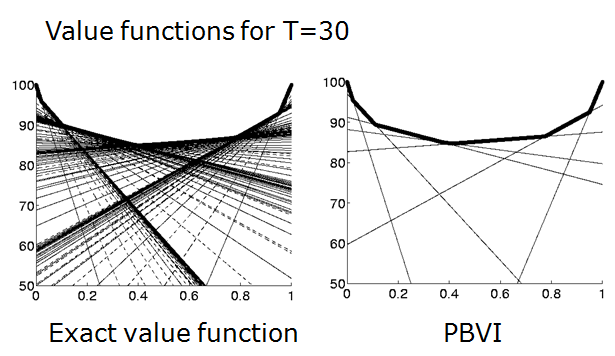}
  \caption{A visual comparison between the exact value function and the approximated function from the PBVI algorithm~\cite{bib:probabilisticrobotics}}\label{fig:pbvi}
\end{figure}
A common feature of offline approaches is that the algorithm returns a policy which defines an action for every possible belief state. Hence, these techniques deal only with small or mid-size problems. Another proposed solution is to use \emph{online} methods which aim to find a good local policy for the current belief state of the agent. Generally, these techniques use approximate offline methods to compute lower and upper bound of the optimal value function. Below, we summarise the most known techniques for online algorithms~\cite{bib:pineau}:
\begin{itemize}
    \item Branch-and-Bound Pruning
    \begin{itemize}
        \item Uses AND-OR tree
        \item Maintains lower and upper values of $Q^{*}(b,u)$ for every belief and action in the tree 
        \item Back propagated
    \end{itemize}
    \item Monte-Carlo Sampling
    \begin{itemize}
        \item Reduces the branching factor at only the belief state reached during the simulation
        \item The simulator may be a black box
    \end{itemize}
    \item Heuristic Search
    \begin{itemize}
        \item Expands only fridge nodes in according with the heuristic 
    \end{itemize}
\end{itemize}

\subsubsection{Exploiting Domain Knowledge in Planning for Uncertain Systems}\label{sec:candido}

Information-feedback control policies, i.e. belief- or sensor-based policies, select an action at any stage in according to data provided at each stage. Since it is impossible to know exactly the future inputs (belief state or sensory data) predicting the evolution of the system may be extremely difficult. In~\cite{bib:candido10} is proposed a planning algorithm which uses domain knowledge about specific problems to reduce the computational costs with proof of concept in the application of a simulated multi-robot firefighting problem and robot navigation problem~\cite{bib:candido11}. The main idea is to define a set of local self-stopping policies which determines the local behaviour of the agent. Each policy has a termination condition that informs the global switching policy when the local policy should no longer be continued. 
\begin{figure}[t]
  \centering
  \includegraphics[width=2in]{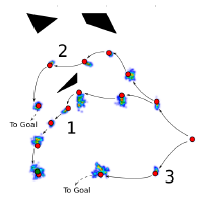}
  \caption{Minimum uncertainty robot navigation using information-based POMDP planning}\label{fig:candido}
\end{figure}
Figure~\ref{fig:candido} shows the robot navigation problem solved by using local information-based control policies. The agent uses two local policies, namely: (1) $\pi_{1}$: minimises the next-stage expected entropy (reduces uncertainty in the robot's state) defined as in equation~\ref{eq:entropy} and (2) $\pi_{2}$: draws the robot towards the goal using the Kullback-Leibler divergence as shown in equation~\ref{eq:kl}.
\begin{equation}\label{eq:entropy}
\mathcal{H}(x)=-\int_{\mathcal{X}}{b_{t}(x)\log_{2}{b_{t}(x)}dx}
\end{equation}
In the example, the agent draws three trajectories in the belief space. The red dots refer to belief states, whilst the blue blurs refer to related uncertainty. The switching policy select the trajectory labeled ``1'' because is affect by less uncertainty than the trajectory labeled ``3'', but reaches the goal faster than the trajectory label ``2''.
\begin{equation}\label{eq:kl}
KL(b^{i}||b^{j})=\int_{\mathcal{X}}{b^{i}(x)\log{\frac{b^{i}(x)}{b^{j}(x)}}dx}
\end{equation}

\subsubsection{Some Applications}

In~\cite{bib:sarsop} a new point-based POMDP algorithm, termed \textbf{SARSOP}, is presented. Here, the key point is to exploit the notion of \emph{optimally belief space} to improve computational efficiency. In short, the authors propose a method to sample points only near the subset of belief points reachable from $b_{0}$ under the optimal sequence of actions. The algorithm iterates over three main functions, namely: SAMPLE, BACKUP and PRUNE. Although the pruning takes up a significant fraction of the total computation time, it guarantees to prune those points that are provably suboptimal.
\begin{figure}[t]
  \centering
  \includegraphics[width=2.5in]{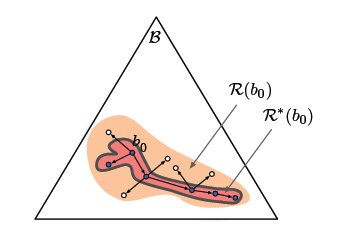}
  \caption{Belief space $\mathcal{B}$, reachable space $\mathcal{R}(b_{0})$, and optimally reachable space $\mathcal{R}^{*}(b_{0})$~\cite{bib:sarsop}}\label{fig:sarsop}
\end{figure}

The proof of concept is applied in a variety of robotic task as well as grasping~\cite{bib:grasp}. From the model point of view, this problem has been approached similarly to coastal navigation: the environment is assumed static and knows, but due to limited sensing capabilities, the agent has partial ignorance of its own state. To reduce the uncertainty, the robot performs compliant guarded moves and always maintains contacts with the surface of the object attempting a stable grasp.
%\begin{figure}[t]
%  \centering
%  \includegraphics[width=2.0in]{img/blocks_world.png}
%  \caption{The robot has successfully put objects with green and red labels into separate piles using probabilistic inference for planning and control~\cite{bib:toussaint}}\label{fig:toussaint}
%\end{figure}

A typical scenario that has a great tradition in the Artificial Intelligence community is the \emph{blocks-world} scenario. Here, the agent has to manipulate objects (blocks) to achieve a specific configuration of the world. In~\cite{bib:toussaint} a real blocks-world has been realized using a 14DOF Schunk arm and hand with tactile sensors and a stereo camera, the goal is to manipulate a set of objects on the table in a goal-oriented way. The key aspect of this work a symbolic representation of states and actions which leads to high-level rule-based planning in relational domains. The work demonstrates the flexibility of approximate inference methods for control and trajectory optimization on the motor level as well as high-level planning in an integrated real world, although sensors uncertainty is not taken into account (i.e. accidentally pushing objects off pile).

%---------------------------------------------------------------------------------------------------------------------
\subsection{Hybrid models}\label{sec:hybrid}

This section describes hybrid models on which path planners for high-dimensional spaces are extended to cope with partial ignorance or approximate POMDP approaches are applied to high-dimensional continuous spaces. 

%---------------------------------------------------------------------------------------------------------------------
\subsubsection{RRT for path planning in continuous space with uncertainty}\label{sec:prrt}

Recently, modified RRT-based algorithms were developed to cope with various kind of uncertainty in the search space. 

In~\cite{bib:melchior}, the \textbf{Particle RRT (pRRT)} algorithm explicitly considers uncertainty in its domain, in a similar way to a particle filter. Each extension to the search tree is treated as a stochastic process and is simulated multiple times. The behavior of the robot can be characterized based on the specified uncertainty in the environment and guarantees can be made as to the performance under this uncertainty. Extensions to the search tree, and therefore entire paths, may be chosen based on the expected probability of successful execution. The benefit of this algorithm is demonstrated in the simulation of a rover operating in rough terrain with unknown coefficients of friction.
\begin{figure}[t]
  \centering
  \includegraphics[width=2.5in]{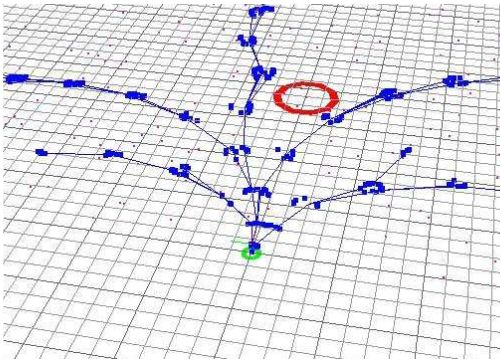}
  \caption{A pRRT tree with several particles at each node~\cite{bib:melchior}.}\label{fig:rrt}
\end{figure}

%---------------------------------------------------------------------------------------------------------------------
\subsubsection{POMDPs in continuous spaces}\label{sec:mcpomdp}

The work shown in~\cite{bib:thrun} proposes an approximate approach, the \textbf{Monte Carlo POMDP (MC-POMDP)}
algorithm, which can accommodate real valued spaces and models. The author is interested in POMDPs with continuous state and action spaces in order to generalize the model for a large number of real-world problems that are continuous in nature. 
The central idea is to use Monte Carlo sampling for belief representation and propagation, while reinforcement learning in belief space is employed to learn value functions, using a sample-based version of nearest neighbour for generalization. Empirical results illustrate that this approach finds close-to-optimal solutions efficiently. Furthermore, in according with the author, initial experimental results demonstrate that this approach is applicable to real-valued domains, and that it yields good performance results in environments that are, by POMDP standards, relatively large.

In~\cite{bib:silver} is presented the state-of-the-art for online planning in large POMDP, termed \textbf{Partially Observable Monte-Carlo Planning (POMCP)}. The algorithm uses sampling techniques to break the curse of dimensionality combining a Monte-Carlo update of the agent's belief with a Monte-Carlo search tree (MCST) form the current belief state. Moreover, the work shows that only a black box simulator of the POMDP model is required rather than explicit probability distribution over states, actions and observations. Unlike precedent methods, this technique provides scalable performance in a variety of challenging problems such as $10\times10$ battleship and partially observable PacMan with approximately $10^{18}$ and $10^{56}$ states respectively. POMCP consist of a \emph{Upper Confidence Bound applied to Tree (UTC)} search~\cite{bib:bandits} that selects actions at each time-step and a particle filter that updates the agent's belief. In practice, each state of the MCST is viewed as a multi-armed bandit and actions are chosen by using the UCB1 algorithm~\cite{bib:bandits}. It is important to note that the UCT algorithm has been extended to partially observable environments by using a search tree of histories instead of states.
However, the POMCP algorithm is based on rollout policy to update the agent's belief which involves look-ahead strategies and therefore prediction on the most likely system evolutions.

%---------------------------------------------------------------------------------------------------------------------
% 	EXISTING WORK ON POMDP & MAREK'S THESIS
%---------------------------------------------------------------------------------------------------------------------
%---------------------------------------------------------------------------------------------------------------------
\subsubsection{Forward models}\label{sec:forwardmodel}

When the hand is in contact with the object to be manipulated, in order to achieve the task, either grasping or manipulation, we necessitate the ability to predict the behaviour of the object under the manipulative actions. For this reason, it is necessary to have a forward model which describes how our actions affect the environment. In addition, the choice of the planning formalism is strictly correlated with the forward model chosen. Two are the possible options: (1) physics engines and (2) learned predictors.

In \textbf{physic engines}, objects are defined by geometric primitives and their motions are predicted in terms of rigid body transformations using the law of physics. Physics Engine subsequently detect all collisions as sets of contacts and modify the movement of simulated bodies using contact resolution methods. Unfortunately, none of these methods are without drawbacks~\cite{bib:kopicki_2010}. Friction and restitution are particularly difficult to model and frequently lead to situations which violate the law of energy conservation. Furthermore, the order in which contacts are resolved is critical and has a great influence on the predicted motion. On the other hand, continuous methods are relatively insensitive to the contact resolution order since they explicitly handle deformation during a single contact. These models, however, are expensive and appropriate parameters of the model can be difficult to obtain in practice. 

\textbf{Learned predictors}, instead, are able to encode physics information without explicitly representing physics knowledge. In the work presented in~\cite{bib:kopicki_2010}, the author shows that various geometric relations between parts of objects can be represented as statistically independent \emph{shape/contact~experts} distributions, and when used in \emph{products of experts} allow us to generalize over shape and applied actions, as well as to learn effectively in high dimensional space.
In other words, the study in~\cite{bib:kopicki_2010} is about predicting what can happen to objects when they are manipulated by an agent, for example, a robot. Although, in this study, the author considers only simple pushing manipulation by a robot, the findings are more generally applicable to predicting more complex interactions.
Prediction is already used in robotic manipulation, in particular when it involves planning
and interaction with the real world. As the real world is governed by laws of physics,
 most previous robotic approaches use either physics simulators or other kinds of
physics-derived parametric models. This has led some researchers to
suggest the abandonment of analytic approaches in some cases: “Clearly analytical solutions to
the forward dynamics problem are impossible except in the simplest of cases, so simulation-based
solutions are the only option”~\cite{bib:cappellieri_icra_2006}.
\begin{figure}[t]
  \centering
  \includegraphics[width=4in]{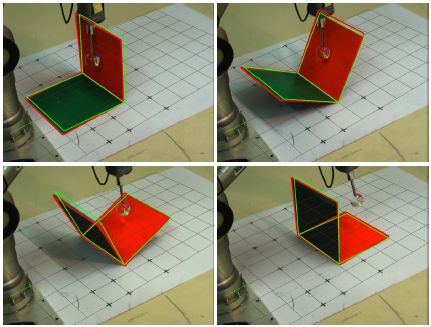}
  \caption{The example shows the interaction between a 5-axis Katana robotic manipulator and an L-shape object, called a polyflap. The green wire frame denotes the prediction whilst the red wire frame denotes the visual tracking~\cite{bib:kopicki_2010}.}\label{fig:marek}
\end{figure}
Learning of forward models is one of the most promising alternatives which could avoid
many of the aforementioned problems since it does not need to refer to any fixed model of the
world, and thus avoids the limitations of such models.

In this study, the author explores alternative approaches to using physics engines, including
learning methods. Specifically: (1) how forward models can be learned with a high accuracy and generalised to
previously unencountered objects and actions; and (2) explore a simplified physics approach which can be combined with the prediction learning
approach.
The contributions of the thesis are numerous:
\begin{enumerate}
\item it poses the learning to predict problem as a problem of probability density estimation. The author contrasts it with a regression formulation and show that density estimation has some advantages: (i) it enables predictions with multimodal
outcomes, (ii) it can produce compromise predictions for multiple combined predictors;
\item it shows how in density estimation we can employ a product of experts architecture to carry
out learning and prediction;
\item it shows that a product of experts architecture can produce generalization with respect to: (i) push direction, (ii) object shape. The author explores various alternative products of experts for encoding the object shape and shows that the best product of experts encodes
constraints, for example, pairs of surfaces or contacts between interacting objects.
\end{enumerate}

%---------------------------------------------------------------------------------------------------------------------
% 	ADAPTIVE CONTROL
%---------------------------------------------------------------------------------------------------------------------
\section{Controller for RRTs}\label{sec:adaptivecontrol}

In this section we expose in detail our approach to solve the problem of grasping and manipulation by an autonomous intelligent agent. The section~\ref{sec:solution} mainly introduces two frameworks, namely: (1) Rapidly-exploring Random Trees for planning  in high-dimensional continuous spaces and (2) Partially Observable Markov Decision Processes for planning with uncertainty. Although both frameworks have been extended in different ways to cope with imperfect information and high-dimensional spaces respectively, the space in between them has not been explored yet by A.I. researchers. Nevertheless, the proposed ideas in this thesis fit entirely the last research directions which attempt to break down the curse of dimensionality of planning in real, complex environments using Monte-Carlo techniques to reduce significantly the complexity and approximate the solution with arbitrary precision.
It seems to us that the properties of the two aforementioned frameworks may be combined together to obtain a new hybrid model which is able to explore high-dimensional environments in autonomous way, guaranteeing a Monte-Carlo search biased towards the largest Voronoi area in the solution space in order to find a possible goal state sampling randomly the high-dimensional continuous state space; and contemporaneously deriving a sensor-based controller which samples multiple actions from a high-dimensional continuous action space in according to a given probability density distribution. It is important to underline that the adaptive controller is defined to have a high-dimensional continuous state space as well, but since we assume to derive the controller from a given initial position, we simply focus on those (finite number of) states that are reachable from the initial configuration in according to the (finite number of) sampled actions. Additionally, the evolution of the system between the reachable states is deterministic in accordance to the given policy. In fact, the controller selects only those trajectories which maximise (or minimise) the reward (cost) function. 

RRTs are popular techniques for path planning with kinodynamic constraints in high-dimensional spaces. Unlike other methods (e.g. PRMs~\cite{bib:kavraki_1996}) they can be intuitively considered as a Monte-Carlo way of biasing search into the largest Voronoi regions. It is also possible extend the basic RRT algorithm for taking into consideration uncertainty related to action's outcome, as introduced in section~\ref{sec:prrt}. The main problem in the application of RRTs to robotic manipulations concerns the challenging computation of an inverse model. In fact, the uncertainty of finding a desired configuration, which solves all the problem constraints, is achieved iteratively by growing randomly an exploring tree, but extending the tree requires the solution of the underlying optimal control problem for which a deterministic solution may not exist~\cite{bib:madani}. As for predicting optimal policies which maximize the sum of rewards in infinite-horizon POMDPs, the problem of approximating an inverse model for robotic manipulation may require a decision-theoretic formalization where there is no direct access to the state of the environment (noisy observations) and the model evolves in accordance to a stochastic transition function, which maps a previous belief states and an actions into a new belief state.\\
In short, the main question this thesis will attempt to solve can be stated as:
\begin{quote}
given a current pose (position and orientation) of the object to be manipulated and a target pose, how to compute the optimal action (or sequence of actions) to move the object to a desired configuration?
\end{quote}
Alternatively, it may be useful to state the problem in a different way, taking explicitly into consideration some technical details concerning our specific scenario.  Our aim is to formalize the problem mentioned above in accordance with the mathematical formulation of section~\ref{sec:pomdpframework}. However, before we can do this, it is necessary to introduce some concepts of robot kinematics. In the following subsections, we offer a brief discussion of the kinematics of robotic manipulations, and then describe in a formal way the problem we wish to solve.

\subsection{Robot Kinematics}

Robot kinematics is the study of the motion of a robot in terms of purely geomerical contraints( i.e. leaving aside considerations of the relationship between forces, inertias and accelerations). A serial robot arm can be modeled as an open chain of links connected by (typically rotatable) joints. In a kinematic analysis, the position, velocity and acceleration of points on an open chain of links are computed without considering the forces that cause the motion. Instead,  typically, the positions and velocities of an end effector are related to angles and angular velocities at the robot’s joints. Robot kinematics also deals with aspects of redundancy, collision avoidance and singularity avoidance. We mainly have two types of robotic kinematics, namely: (1) \emph{forward} or \emph{direct kinematics} and (2) \emph{inverse kinematics}. The former determines the position of the robot's end effector with respect to a global frame of reference, given the length of each link and the angle of each joint. The latter, on the other hand, computes the angle which each joint should assume to deliver the robot's end effector to a desired global position, given the length of each link.

\begin{figure}[t]
  \centering
  \includegraphics[width=3in]{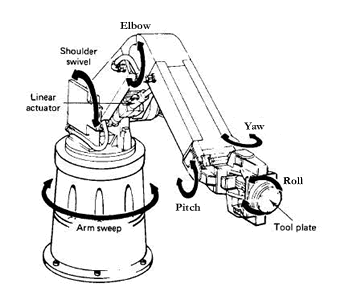}
  \caption{6-DOF jointed arm robot. The basic three rotary joints enable arm swap, shoulder swivel and elbow rotations. In addition, three revolute joints allow the robot to point in many directions}\label{fig:articulatedarm}
\end{figure}

With reference to the robotic manipulation and grasping, let $\mathcal{C}$ be the set of all possible configurations of the object to be manipulated with respect to a global frame of reference $\mathcal{O}$. Any point $p\in\mathcal{C}$ is expressed as $p=[R_{\mathcal{O}}t_{\mathcal{O}}^{T}]$, where $R_{\mathcal{O}}\in\Re^{3\times3}$ is a rotation matrix and $t_{\mathcal{O}}\in\Re^{3}$ is a transition vector, both over the $x$, $y$ and $z$ axis with respect to $\mathcal{O}$. Hereafter we also refer to this set as the \emph{working space}. The working space should not be confused with the \emph{configuration} or \emph{joint space} $\mathcal{J}$, of a robotic device which defines the set of all possible configurations that the robot can assume but express in joint angles. To make this point clearer, figure~\ref{fig:articulatedarm} shows all the joints of a 6-DOF articulated or jointed arm robot. Any configuration of the robot can be expressed by a 6-dimensional vector which specifies the angles each joint should assume.

\subsection{Global Planner}

\subsection{Robotic manipulation and grasping as POMDPs}

\begin{figure}[t]
  \centering
  \includegraphics[width=3in]{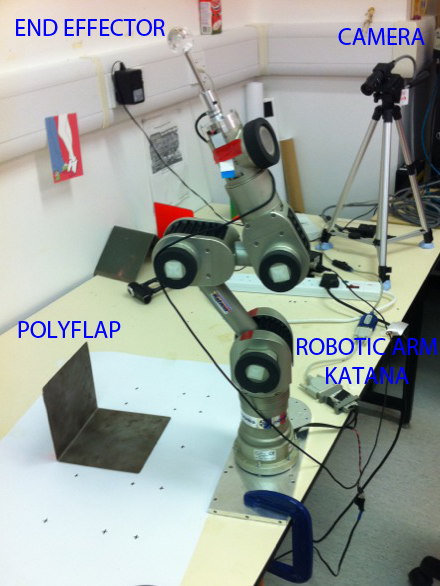}
  \caption{The real pushing scenario with the Katana 6M180, its spheric end effector, a polyflap and the camera which captures noisy observation of the environment.}\label{fig:scpushing}
\end{figure}

Let $\mathcal{M}=\langle\mathcal{X},\mathcal{U},\mathcal{Y},\tau,\psi,\rho\rangle$ be a POMDP model, where:
\begin{itemize}
\item $\mathcal{X}$ defines configurations of the environment express as the pose of the object and of the agent in working space $\mathcal{C}$;
\item $\mathcal{U}$ defines movements of the agent in joint configuration space $\mathcal{J}$;
\item $\mathcal{Y}$ defines observations of the object's pose;
\item $\tau\colon\mathcal{X}\times\mathcal{U}\rightarrow\Pi(\mathcal{X})$ defines the state transition function;
\item $\psi\colon\mathcal{X}\times\mathcal{U}\rightarrow\Pi(\mathcal{Y})$ defines the observation function;
\item $\rho\colon\mathcal{X}\times\mathcal{U}\rightarrow\Re$ defines the cost function to minimise;
\end{itemize}
Figure~\ref{fig:scpushing} shows the real pushing scenario with the Katana 6M180 arm, its spheric end effector which is used as a robotic finger and the L-shape object termed polyflap which will be manipulated by the robot. All our knowledge about the environment configuration is inferred by the camera which captures the position and orientation of the L-shape object, termed \emph{polyflap}, with reference to a global frame at each time step $t$.

Such models are computationally intractable mainly for the following reasons:
\begin{enumerate}
\item high-dimensional continuous state space;
\item non-linear transition and observation functions;
\end{enumerate}
The state space $\mathcal{X}$ requires a high representational cost which depends directly on the complexity of the agent. In other words, the greater the number of the robot’s joints, the higher the dimension of $\mathcal{X}$ will be. As mentioned in section~\ref{sec:problem}, the principal uncertainty is in the configuration of the robot and the state of the objects in the world, that is formally represented by the set $\mathcal{X}$. Furthermore, the high-dimensional state space also means that many parameters must be set in order to properly describe  the system evolution. Additionally, uncertainty related to the outcomes of robot actions, make it impossible to define analytical or deterministic transition and observation functions.

Our proposed solution can be summarised as follows:
\begin{enumerate}
\item use of a black box simulator for POMDPs instead of an analytic model;
\item deterministically define the system evolution according to a set of local information-based control policies;
\end{enumerate}
More specifically, we plan to use a simulated physic environment (see section~\ref{sec:forwardmodel}) as a black box simulator for POMDPs rather than an explicit probability distribution over states, actions and observation spaces. Then, according to the authors of~\cite{bib:silver}, we only need to define a reward or cost function in order to accurately drive  the system towards an optimal solution. 
We propose to use a switching control policy as described in section~\ref{sec:candido}. The switching policy should be considered the high-level planner, whilst at the low level we define a set of local information-based policy which specifies the system evolution. Each local policy considers only one cost (or equivalently reward) function and at any time only one local policy can be selected by the high-level planner. Hence the evolution of the system is deterministically computed drawing a finite number of reachable belief points from the current belief point and selecting the one which minimises (maximises) the current cost (reward) function.

\subsection{Implementation} 

Appendix~\ref{sec:pseudocoderrt} presents the original RRT algorithm (algorithm~\ref{alg:buildrrt}) as defined in~\cite{bib:lavalle_2001}, whilst the algorithms~\ref{alg:randomstate}~\ref{alg:selectaction}~\ref{alg:newstate} define our first attempt at applying RRTs to the problem of pushing manipulation. As described in RMSG Report 2, our principal modification to the original algorithm lies in the algorithm~\ref{alg:selectaction} which implements a simple heuristic-driven substitute for an inverse model. In short, we select an action to extend  nodes of the tree by moving the finger along the straight line which connects the central of mass of the object to the current target point, which we hope to add to the tree. This implementation is simplistic since it does not encode any information about the orientation or rotation motions of the object, and ignores certain aspects of the object's shape. Nevertheless, it showed good results  for cubic shaped objects, but extends poorly to other objects, e.g. the polyflap object which can tip or topple as well as slide.

In order to improve the simple, early version of our algorithm, we propose to re-implement algorithm~\ref{alg:selectaction} as an adaptive controller. Like the POMCP algorithm (see section~\ref{sec:mcpomdp}), we define the stochastic embedded system as a black box simulator for the POMDP model and the agent's belief is updated by using a Monte-Carlo technique. In a different way, we do not use rollout techniques to explore trajectories in the solution space. Instead we design some sensor-feedback policies, as define in section~\ref{sec:candido}, to directly explore  the optimal area of the solution space.
Algorithm~\ref{alg:buildporrt} shows in pseudo-code our modified implementation of the basic RRT algorithm. The algorithm iteratively constructs an RRT $T$, rooted in the beginning state configuration $x_{init}$, until the goal configuration is reached with a maximum of $K$ vertices. 

\begin{algorithm}                      
\caption{BUILD\_PORRT}          
\label{alg:buildporrt}                           
\begin{algorithmic}                    
\REQUIRE $x_{init}, K, \epsilon$
\ENSURE $T$
\STATE $T.$init($x_{init}$)
\FOR{$k=1\to K$}
  \STATE $x_{rand}\Leftarrow\,$RANDOM\_STATE()
  \STATE $x_{near}\Leftarrow\,$NEAREST\_neighbour($x_{rand}, T$)
  \STATE $\pi\Leftarrow\,$SELECT\_CONTROL\_POLICY($x_{near}, x_{rand}, K, \epsilon$)
  \STATE $x_{new}\Leftarrow\,$NEW\_STATE($x_{near}, \pi$)
  \STATE $T.$add\_vertex($x_{new}$)
  \STATE $T.$add\_edge($x_{near},x_{new},\pi$)
  \STATE control stop condition $x_{new}$ is the goal
\ENDFOR
\end{algorithmic}
\end{algorithm}

\begin{algorithm}                      
\caption{SELECT\_CONTROL\_POLICY}          
\label{alg:selectcontrolpolicy}                           
\begin{algorithmic}                    
\REQUIRE $x_{near}, x_{rand},\, K,\, \epsilon$
\ENSURE $x_{best}$
\STATE $x_{new}\Leftarrow\,x_{near}$
\STATE $\pi\Leftarrow\,\emptyset$
\FOR{$k=1\to K$}
  	\STATE $inputs\Leftarrow\,$SAMPLE\_RANDOM\_ACTIONS($x_{new})$
  	\STATE $x_{best},u_{best}\Leftarrow\,$POMDP\_SIMULATOR($x_{new},inputs$)
	\STATE $\zeta_{best}\Leftarrow\,$COST\_FUNCTION($x_{best},x_{rand}$)
	\STATE $\pi$.add\_element($x_{best},u_{best},\zeta_{best}$)
	\STATE control stop condition $\zeta_{best}<\epsilon$
	\STATE $x_{new}\Leftarrow\,x_{best}$
\ENDFOR
\end{algorithmic}
\end{algorithm}

Algorithm~\ref{alg:selectcontrolpolicy} shows the pseudo-code of the adaptive controller. So far, the proposed algorithm has been implemented only in terms of a single pushing scenario and briefly can be expressed as follows: at time $k$, the RRT algorithm randomly selects  a point in the continuous working space, defined as rotation and translation with respect to a reference frame. At this stage, we are able to compute a metric distance, which takes into account either angular or linear distance, between the current pose of the object and the desired one. The agent then considers some random actions (sampled according to some heuristic for the time being) to make the object move. Each such action is trialled in a physics simulator to predict what it’s outcomes might be. If the agent realises that an action's outcome is increasing the metric distance, the action is immediately discarded. On the other hand, if the action's outcome produces any improvement, the action is continued until a local minima of the cost function is reached. A fixed number of actions are explored at each iteration, and only the one which minimises the cost function is executed by the controller at time $k$, thus extending the RRT tree as close as possible towards the next new node requested by the RRT algorithm.

\section{Evaluation of the work}\label{sec:evaluation}

This section concerns with how we will design experiments to measure the performance of our system. We plan to analyse the system performance in terms of both quantitative and qualitative metrics. The former defines a measurement of success for task-oriented behaviours which specifies  the accuracy with which a system is able to manipulate  object in order to achieve a desired goal configuration. The latter will require a protocol for deciding whether or not the object pose achieved by the system qualitatively matches that of the goal, e.g. is a box object toppled so that it rests on the correct desired face, or not? Note that designing these performance measurements is important for two different and very distinct reasons as state below:
\begin{itemize}
\item they are necessary for evaluating the global performance of the system;
\item they also provide cost functions that can be used to train or guide our actual manipulation algorithm while it is working.
\end{itemize} 

In order to generate performance measurements, we intend to carry out experiments both in a simulation environments and also with real experiments on a real robot. Both of these kinds of experiments are important. Simulated data are necessary because they provide perfectly known ground-truth data for performance evaluations in a fully measurable environment. Moreover, simulation allows us to carry out a very large number and a variety of experiments. On the other hand, real world experiments are essential to properly demonstrate that a robotic system works, when subject to the large, varied and very complex kinds of noise and uncertainty of the real world actuators, sensors and objects. However, it is very difficult and time-consuming to set up real experiments, so it is usually only possible to carry out a relatively small number of tests on a small set of test objects. Furthermore, it is impossible to collect good ground truth data with real robotic manipulation experiments, i.e. the poses of the manipulated objects are captured by a vision system which itself is prone to significant errors. Hence, performance measures of success will themselves be prone to errors. For these reasons, it is important to provide a combination of both real and simulated  experiments to properly evaluate the performance of our system. 

\begin{figure}[t]
  \centering
  \includegraphics[width=2.5in]{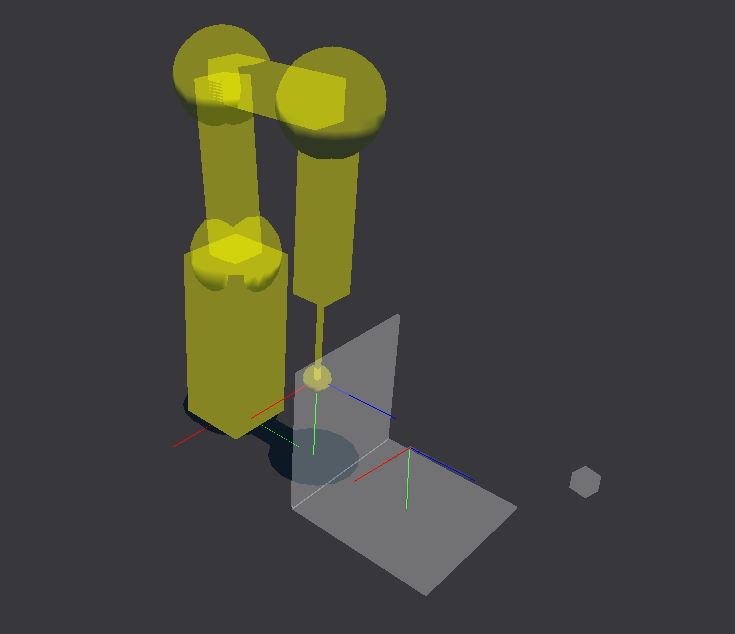}
  \caption{Golem-based virtual environment with a 5-axis Katana arm and a polyflap for the pushing scenario}\label{fig:golem}
\end{figure}

Figure~\ref{fig:golem} shows a picture of the simulation environment in which we will carry out pushing manipulation experiments. An arm with a simple, rigid finger, is able to apply a variety of pushes to a variety of simulated objects, whose resulting motions can be modelled with physics simulation software. The robot will be tasked with planning a series of pushes that move an object as close as possible to a desired goal position. Once the robotic actions are complete, quantitative and qualitative distance metrics can be used to evaluate the success of the robotic actions. This procedure can be repeated for a large number of different objects with randomised starting and goal states.

Figure~\ref{fig:scpushing} shows the real experimental setup in the lab, which we will use for a real-world implementation of these experiments, once simulation experiments have been completed.

The research presented in this proposal is predicated on the assumption that the problem of object manipulation from a robotic hand should be modelled as a stochastic embedded system. It has been mentioned that control theorists have developed a rich framework for controlling stochastic dynamic systems by using adaptive controllers. Our aim is to use and possibly extend such frameworks with proof of concept in both applications to planning in the pushing and grasping scenarios. 
First, we will arrange a virtual scenario based on Golem framework~\cite{bib:kopicki_2010} using a physics-based forward model on where our algorithm will be tested (see figure~\ref{fig:golem}). In this case, we can assume that configurations of the environment are static and fully known, and the only uncertainty is related to how the environment changes in response to the agent's actions. Eventually, the use of a physics engine as a forward model will be replaced with a learned predictor (described in~\ref{sec:forwardmodel}) and performance using the two alternative forwards models will be compared. As already mentioned, in the pushing scenario the task is to design an autonomous controller to enable the robot to move an object towards a goal position, using single finger pushing actions alone. Since the virtual environment is fully observable, we can accurately compute a distance metric  $\zeta$ (between present object pose and the desired goal pose) in terms of angular and linear displacements. This metric is useful because it can be used to quantify the performance of the system, i.e. over a series of experiments it is possible to quantify how well the proposed manipulation system is able to move objects near to a desired goal position. Equation~\ref{eq:metric} shows how to compute the metric value: $p_{1},p_{2}\in\mathcal{C}$ are two points in the working space, the variables $\alpha,\beta\in[0,1]$ are constant scalar values which sum up to 1, $quat(\cdotp)$ is the quaternion operator which transforms a point express in $\Re^{3\times4}$ into $\Re^{4}$, the operators $||\cdotp||_{q}$ and $||\cdotp||_{l}$ are the quaternion metric and the Euclidian distance, respectively. This value is interpreted by the local policy as a cost function to be minimised, as described in section~\ref{sec:adaptivecontrol}.
\begin{equation}\label{eq:metric}
\zeta=\alpha||quat(p_{2})-quat(p_{1})||_{q}+\beta||p_{2}-p_{1}||_{l}
\end{equation}

After we have experimented with the proposed system in simulation, we plan to implement the system on a real robot arm, in a real scenario where it is no longer possible to assume a static and fully known environment. In practice, the system will observe the scene by using noisy sensors as, e.g. a camera and vision algorithm which exhibits significant errors when tracking manipulated objects. Therefore, our model will be tested in a real much more uncertain world than the initial simulation studies. 
%So far, we need to define only one local policy which draws the system towards the desired pose of the object, because we assume the observations from the virtual environment match exactly the working scenario and we therefore do not consider uncertainty over the set of observations. 
In order to cope with noisy observations in the real environment we will add some additional local policies which constrain the robot to move in safe ways, avoiding collisions and smoothly approaching contacts with objects. These local policies could also be tested in simulation experiments where we deliberately introduce varying amounts of noise. As already mentioned, simulation is a powerful tool to carry out a large number of experiments with different variety of settings. One way of doing that is to model vision by taking the perfect pose of the object from the simulation and then adding in an artificial error (e.g. Gaussian noise with various different size standard deviation) before passing the new, uncertain position to our control algorithm. When observations are no longer certain, the agent needs the ability to infer specific information from the noise  in order to complete the task, i.e. before start pushing the object the agent needs to know whether or not its end effector is in contact with it.
With this approach we are able to explicitly model uncertainty in simulation as well as in the real world, and define local policies which may converge the system evolution towards those states where our confidence is maximised in order to achieve safely the task.  

\begin{figure}[t]
  \centering
  \includegraphics[width=3in]{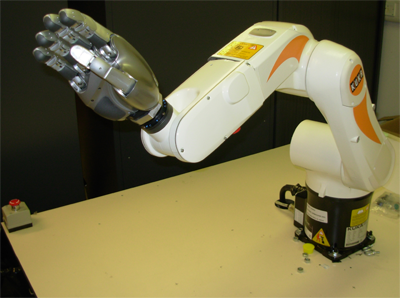}
  \caption{The real grasping scenario with the KUKA arm KR 5 sixx R850 6DOF and the five-finger DRL-Hand II.}\label{fig:scgrasping}
\end{figure}

The future work would be to extend this approach to multiple fingers manipulation and grasping with a robotic hand. We might involve in-hand manipulation experiments with the grasping setup in the lab (see figure~\ref{fig:scgrasping}) as well as more simple experiments where the Katana single finger moves an object that is simultaneously supported by several other stationary ``fingers''. In the last months, we have developed a simulation environment for the arm and the hand based on openRAVE (Open Robotics Automation Virtual Environment). This framework allows us to simulate and analyse of kinematic and geometric information related to motion planning and can be easily used with the most popular physics engines. However, our recent work has shown that it is very hard to simulate the physics of multiple contacting fingers so that we may find that real experiments become much more important as we move from one finger to a multi-fingered hand.

%---------------------------------------------------------------------------------------------------------------------
\section{Related work: Flexible automation of micro and meso-scale manipulation}\label{lab:cappelleri}

\begin{figure}[t]
  \centering
  \includegraphics[width=3in]{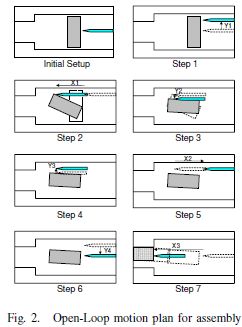}
  \caption{Open-loop motion plan for assembly~\cite{bib:cappellieri_icra_2006}.}\label{fig:cappelleri}
\end{figure}
The study presented in~\cite{bib:cappellieri_icra_2006} describes a test-bed for planar micro
manipulation tasks and a framework for planning based on quasi-static
models of mechanical systems with frictional contacts. It
shows how planar peg-in-the-hole assembly tasks can be designed
using randomized motion planning techniques with Mason’s
models for quasi-static manipulation~\cite{bib:masonthesis}.

The goal is to use simulation and motion
planning tools to design open loop manipulation plans that rely
only on an estimate of initial position and orientation. An example is shown in Figure~\ref{fig:cappelleri}. They use
Mason’s quasi-static models for manipulation of planar parts
with surface friction. A method for 3-D simulation is adapted to solve the “2.5-dimensional” problem
with surface friction. An application of the Rapidly Exploring
Random Tree (RRT) algorithm with modifications for
dynamic systems is used to solve the peg-in-hole insertion task.

%------------------------------------------------------------------------------------------------------------------------------------------------------------------------------------------------
%	FUTURE WORK
%------------------------------------------------------------------------------------------------------------------------------------------------------------------------------------------------
\section{Plan for future work}

In this section, we present a timetable for the tasks we aim to finish over the next two years, culminated in the production of the thesis. Our immediate goal is to write up as a paper the results obtained on the pushing scenario.

\vspace{5mm}
\begin{tabular}{ l l }
July - August & Implementation of an adaptive controller for the pushing scenario\\
& \quad - using the Golem simulator (2 weeks) \\
& \quad - run experiments on the real pushing scenario (2 weeks) \\
& \quad - write the paper (3 weeks) \\
16th September & Deadline submission paper for ICRA-12 \\
November & Deadline submission paper for ACM-12 workshop \\
December - February 2012 & Carrying out a series of experiments to collect better \\ & performance
evaluationdata set \\
&  Analysis of results and write up as a paper\\
23rd February 2012 & Preparing the RSMG Report 4 \\
& \quad - Literature review (1 weeks) \\
& \quad - Writing (2 weeks) \\
& \quad - Review and correction (1 weeks) \\
23rd March 2012 & RSMG Report 4 (GRS1A) submission \\
2nd May 2012 & RSMG meeting deadline\\
May - September 2012 & Extend the framework to a multiple fingers manipulation\\
& Implementation on real robot \\
28th September 2012 & RSMG Report 5 (GRS1B) submission\\
7th November 2012 & RSMG meeting deadline\\
December - March 2013 & Implementation of an adaptive controller for the grasping scenario\\
 & End of the GeRT project\\
29th March 2013 & RSMG Report 6 (GRS1A) submission\\
1th May 2013 & RSMG meeting deadline\\
May 2013 & Start writing the thesis\\
4th October 2013 & RSMG Report 7 (GRS1B) submission\\
6th November 2013 & RSMG meeting deadline

\end{tabular}

%------------------------------------------------------------------------------------------------------------------------------------------------------------------------------------------------
%	APPENDIX
%------------------------------------------------------------------------------------------------------------------------------------------------------------------------------------------------
\newpage
\appendix
\section{Appendix: pseudo code for RRT}\label{sec:pseudocoderrt}

For a given state, $x_{init}$, an RRT, $t$, with $K$ vertices is constructed as shown below:\\

\begin{algorithm}                      % enter the algorithm environment
\caption{BUILD\_RRT}          % give the algorithm a caption
\label{alg:buildrrt}                           % and a label for \ref{} commands later in the document
\begin{algorithmic}                    % enter the algorithmic environment
\REQUIRE $x_{init}, K$
\ENSURE $T$
\STATE $T.$init($x_{init}$)
\FOR{$k=1\to K$}
  \STATE $x_{rand}\Leftarrow\,$RANDOM\_STATE()
  \STATE $x_{near}\Leftarrow\,$NEAREST\_NEIGHBOUR($x_{rand}, T$)
  \STATE $u\Leftarrow\,$SELECT\_ACTION($x_{near}, x_{rand}$)
  \STATE $x_{new}\Leftarrow\,$NEW\_STATE($x_{near}, u$)
  \STATE $T.$add\_vertex($x_{new}$)
  \STATE $T.$add\_edge($x_{near},x_{new},u$)
\ENDFOR
\end{algorithmic}
\end{algorithm}

\begin{algorithm}                     
\caption{RANDOM\_STATE}          
\label{alg:randomstate}                          
\begin{algorithmic}                   
\ENSURE $x$
\STATE $x\Leftarrow$LOWER\_STATE
\STATE $m\Leftarrow$STATE\_DIMENSION
\FOR{$k=1\to m$}
  \STATE $r\Leftarrow$RANDOM\_NUMBER()
  \STATE $x[i]\Leftarrow r * (UPPER\_STATE[i] - LOWER\_STATE[i])$
\ENDFOR
\end{algorithmic}
\end{algorithm}

\begin{algorithm}                     
\caption{SELECT\_ACTION}          
\label{alg:selectaction}                          
\begin{algorithmic}                   
\ENSURE $u$
\STATE $u\Leftarrow\,$NORMALIZE($x_{rand} - x_{near}$)
\end{algorithmic}
\end{algorithm}

\begin{algorithm}                     
\caption{NEW\_STATE}          
\label{alg:newstate}                          
\begin{algorithmic}                   
\ENSURE $PHYSX\_CUBE\_POSITION$
\STATE $x_{start}\Leftarrow$PHYSX\_MOVE\_END\_EFFECTOR(-u, $DELTA_{b}$)
\WHILE{$x_{rand} -$ PHYSX\_END\_EFFECTOR\_POSITION $< h$}
  \STATE $x_{end}\Leftarrow$PHYSX\_MOVE\_END\_EFFECTOR(u, $DELTA_{f}$)
\ENDWHILE
\end{algorithmic}
\end{algorithm}

%------------------------------------------------------------------------------------------------------------------------------------------------------------------------------------------------
%	BIBLIOGRAPHY
%------------------------------------------------------------------------------------------------------------------------------------------------------------------------------------------------
\clearpage
\bibliographystyle{plain}
\bibliography{../../bib/grasping_under_uncertainty} 
%\bibliography{../../bib/report3}

\end{document}